\let\Xdocument\document

\documentclass[11pt]{article}
\let\document\Xdocument
\usepackage{mathptmx}          % 设置字体为 Times New Roman
\usepackage{setspace}          % 引入setspace包
\usepackage[a4paper, left=1in, right=1in, top=1in, bottom=1in]{geometry}

\usepackage{amsmath}

\usepackage{graphicx}

\usepackage{caption}
\usepackage{booktabs}
\usepackage{array}
\usepackage[english]{babel}
\usepackage{authblk}  % 推荐用于作者+单位排版
\usepackage{hyperref} % 通讯作者链接支持

\title{Drug classification based on X-ray spectroscopy combined with machine learning}
\date{}  % This removes the date from the title
\author[1,2]{Yongming Li}
\author[1]{Peng Wang}
\author[1]{Bangdong Han\thanks{Corresponding author: \href{mailto:remiliasgarlet@gmail.com}{remiliasgarlet@gmail.com}}}

\affil[1]{Xiamen University}
\affil[2]{Fudan University}

\begin{document}

\maketitle

\begin{abstract}
    The proliferation of new types of drugs necessitates the urgent development of faster and more accurate detection methods. Traditional detection methods have high requirements for instruments and environments, making the operation complex. X-ray absorption spectroscopy, a non-destructive detection technique, offers advantages such as ease of operation, penetrative observation, and strong substance differentiation capabilities, making it well-suited for application in the field of drug detection and identification. In this study, we constructed a classification model using Convolutional Neural Networks (CNN), Support Vector Machines (SVM), and Particle Swarm Optimization (PSO) to classify and identify drugs based on their X-ray spectral profiles.
In the experiments, we selected 14 chemical reagents with chemical formulas similar to drugs as samples. We utilized CNN to extract features from the spectral data of these 14 chemical reagents and used the extracted features to train an SVM model. We also utilized PSO to optimize two critical initial parameters of the SVM.
The experimental results demonstrate that this model achieved higher classification accuracy compared to two other common methods, with a prediction accuracy of 99.14\%. Additionally, the model exhibited fast execution speed, mitigating the drawback of a drastic increase in running time and efficiency reduction that may result from the direct fusion of PSO and SVM.
Therefore, the combined approach of X-ray absorption spectroscopy with CNN, PSO, and SVM provides a rapid, highly accurate, and reliable classification and identification method for the field of drug detection, holding promising prospects for widespread application.

\end{abstract}

\section{Introduction}

In recent years, the types of drugs have been continuously increasing. Besides traditional drug-related crimes, the proportion of new drug-related crimes has gradually risen, which has imposed higher demands on drug detection technologies in terms of sensitivity, accuracy, speed, safety, and other indicators. Currently, both domestically and internationally, methods that combine artificial intelligence techniques with spectroscopy, chromatography, immunoassay, capillary electrophoresis, and ion migration spectrometry are widely used for drug detection. Among them, surface-enhanced Raman spectroscopy (SERS) has been fully recognized for its applications in biomedical research, food safety, and drug detection. SERS is a Raman spectroscopy technique that uses noble metal nanomaterials to enhance the signal of target molecules[1]. As one of the representative machine learning algorithms, support vector machines (SVM) have been applied to classify and recognize the spectra of various substances in many SERS studies, yielding satisfactory classification results. Dong R[2] developed a novel, rapid method to detect and directly read out methamphetamine in human urine using dynamic surface-enhanced Raman spectroscopy and a portable gold nanorod Raman spectrometer along with SVM classification algorithm. Despite the promising results, SERS detection has high requirements on detection instruments and environment, complicated operation, and can only detect the surface of object, which has great limitations on the detection of drug smuggling. X-ray absorption spectroscopy technology has the advantages of low difficulty in operation, penetrating observation and strong ability to distinguish substances. In addition, X-ray absorption spectroscopy has proven effective in detecting various materials, including plastics and different inorganic compounds[3], and biological tissues[4]. Based on the above analysis, this paper uses X-ray absorption spectroscopy and machine learning techniques to classify various drugs and similar substances.

In the process of SVM model training, hyperparameter optimization is a very difficult and critical problem. To optimize the model parameters, researchers proposed the particle swarm optimization (PSO) algorithm, which has excellent optimization performance in solving nonlinear problems[5]. The algorithm has the advantages of simple principle, fewer adjustable parameters[6]. In recent years, it has gained increasing attention. Its core idea is to share the information of individual particles in the group to obtain the optimal solution through iterations. However, as the iterations progress, the diversity of the particles will decrease. If the positions of global and local optima are the same as the positions of particles, the algorithm may fall into local optima after a certain number of iterations, resulting in poor global performance. In order to improve the performance of PSO, the inertial weight which changes with the number of iterations is used to improve the search quality of PSO. Finally, in this study, the improved PSO(IPSO) algorithm is used to optimize the parameters of SVM model.

In previous studies, researchers typically used MLP network architectures to extract image features. However, MLPs encounter issues such as parameter explosion and low training efficiency when dealing with complex images. In contrast, Convolutional neural network (CNN) can effectively address these challenges. CNN is a novel feature extraction method and are widely applied in classification models[7]. CNN reduces the number of parameters through parameter sharing between neurons, thereby lowering the complexity of the model and enabling more effective feature extraction from the data. In this paper, CNN is used to extract features from X-ray absorption spectra. Compared with the direct use of CNN for sample classification, it can effectively avoid the problem of gradient disappearance[8], reduce computational costs of parameter optimization and improve the final classification accuracy.

This paper's main contributions are as follows: (1) Proposing a new method of drug detection for effective drug classification, using collected X-ray absorption spectra, combined with improved machine learning algorithm. (2) Using CNN to extract features from X-ray absorption spectra and combining IPSO and SVM to solve the issue of parameter tuning. This method improves classification accuracy and reduces computational costs. Thus, it offers an effective solution for drug classification using X-ray absorption spectroscopy. 

\section{Datasets}
\subsection{Data Collection}
    The experiment employed an X-ray spectroscopy detection system designed and developed by our laboratory, as shown in Figure 1[9]. The core components of the system include an X-ray tube and an X-ray detector. The X-ray tube is a KYW800 model with positive and negative voltage capabilities, manufactured by Shanghai Keyiwei Electronic Technology Co., Ltd. It has a rated power of 80~W, a maximum anode voltage of 80~kV, and a tungsten (74W) anode target material. The X-ray detector is an X-123 CdTe detector produced by AmpTek, USA, with an energy resolution of 850~eV~(FWHM)~@~122~keV~($^{57}$Co). During the experiment, the X-ray tube voltage was set to 30~kV, the tube current to 10~$\mu$A, and the filament current to 1000~mA. The X-ray detector's acquisition time was 30 seconds.

\begin{figure}[h!]
    \centering
    \includegraphics[width=0.6\textwidth]{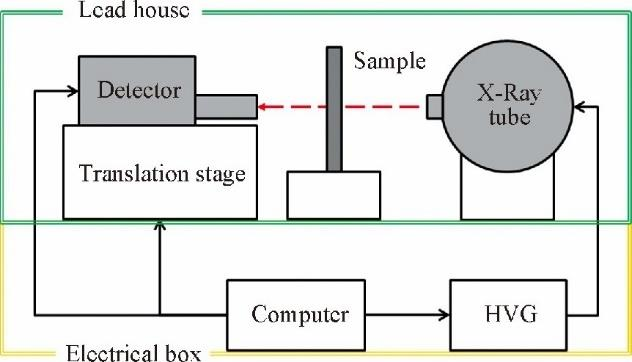}
    \captionsetup{font={small,bf}} 
    \caption{Schematic diagram of X-ray spectral detection system}
    \label{fig1}
\end{figure}

The experimental samples consist of 14 compounds with chemical formulas similar to drugs. Their names, molecular formulas, and corresponding CAS Numbers are listed in Table 1. For convenience in subsequent discussions, the samples are denoted by numbers 1 to 14. These samples were provided by Aceri Technology (Xiamen) Co., Ltd. For the 14 types of experimental samples, 200 valid spectra were recorded for each sample under identical experimental conditions, resulting in a total of 2800 valid spectra.

\subsection{Data Preprocessing}

Data processing includes normalization and shuffling. First, all data are normalized to the range of 0 to 1. Then, 90\% of the processed spectral data are randomly selected as the training set, while the remaining 10\% are used as the testing set. All data processing was performed using Matlab R2021a. The processed spectra are shown in Figure 2. To better observe the differences, the x-axis displays values from channel 20 to 50, and the y-axis is shifted. It can be observed that the spectra are very similar, with the positions of the characteristic peaks being almost identical.

\begin{figure}[h!]
    \centering
    \includegraphics[width=0.7\textwidth]{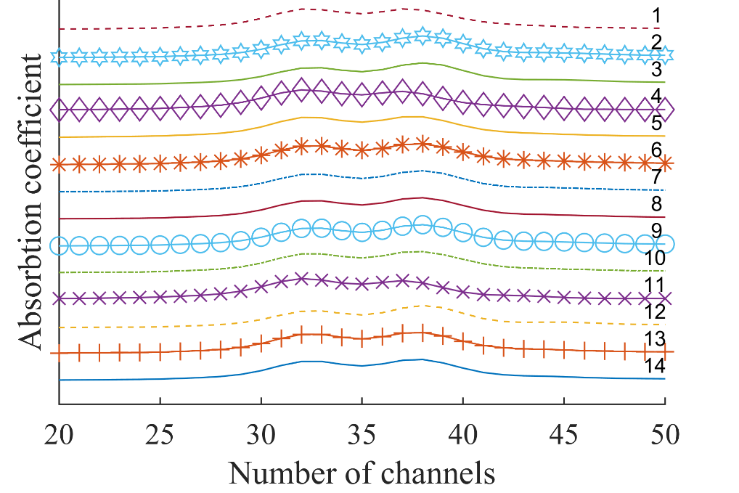}
    \captionsetup{font={small,bf}} 
    \caption{The spectra of 14 samples}
    \label{fig2}
\end{figure}

\begin{table}[h!]
\centering
    \captionsetup{font={small,bf}} 
\caption{CAS Number, molecular formula, and corresponding number of 14 samples}
\setlength{\tabcolsep}{5pt} % Reduce column separation
\renewcommand{\arraystretch}{1.2} % Adjust row height for better readability
\begin{tabular}{@{\hskip1cm}ccc@{\hskip1cm}}
\toprule
\textbf{CAS Number} & \textbf{Molecular Formula} & \textbf{Corresponding Number} \\ 
\midrule
322407-34-1 & C$_{13}$H$_{19}$NO & 1 \\ 
85388-94-4 & C$_{10}$H$_{17}$NO$_{2}$ & 2 \\ 
1836-62-0  & C$_{9}$H$_{13}$NO$_{2}$ & 3 \\ 
3147-75-9  & C$_{20}$H$_{25}$N$_{3}$O & 4 \\ 
103909-86-0 & C$_{13}$H$_{15}$NO$_{2}$ & 5 \\ 
91-68-9    & C$_{10}$H$_{15}$NO & 6 \\ 
86604-78-6 & C$_{9}$H$_{13}$NO$_{2}$ & 7 \\ 
104-13-2   & C$_{10}$H$_{15}$N & 8 \\ 
99-97-8    & C$_{9}$H$_{13}$N & 9 \\ 
1126-78-9  & C$_{10}$H$_{15}$N & 10 \\ 
57-83-0    & C$_{21}$H$_{30}$O$_{2}$ & 11 \\ 
46917-07-1 & C$_{15}$H$_{20}$ClNO$_{2}$ & 12 \\ 
1199-46-8  & C$_{10}$H$_{15}$NO & 13 \\ 
30950-27-7 & C$_{10}$H$_{15}$NO & 14 \\ 
\bottomrule
\end{tabular}
\end{table}

\section{Analysis Methods}
\subsection{Feature Extractor: Convolutional Neural Network}

Convolutional Neural Networks (CNNs), as a deep learning method, are widely applied in recognition, classification, and prediction tasks. The CNN model designed in this paper is based on LeNet-5, a classic CNN architecture used for handwritten Arabic digit recognition. 

The operation of a CNN involves two important processes: the forward propagation process and the backward feedback process. In the forward propagation phase, the entire spectrum from the training set is used as input to the network, and the hidden layers extract and process features from the input. In this phase, the information flows gradually from the input layer to the output layer. The second phase is the backward propagation phase. The network computes the error between the predicted class values for all input samples and the corresponding true outputs, which serves as the target optimization function. The weights in the hidden layers are adjusted based on the error minimization method. The network model uses mini-batch gradient descent to minimize the loss function, adjusting the weight parameters layer by layer in the network, and improves the network's accuracy through frequent iterative training cycles, until the network reaches a high recognition accuracy.

The data extraction part of the model includes two convolutional layers and two pooling layers. Through the fully connected layer, 84 features are obtained, and the dataset is divided into 14 categories based on these 84 features. This paper first discusses the classifier under the traditional architecture: the classification performance of Softmax, with the loss function being cross-entropy loss. The CNN model architecture is shown in Figure 3. The convolution kernel size of the first convolutional layer is set to 3×1, with a stride of 2 and 10 filters; the convolution kernel size of the second convolutional layer is 3×1, with a stride of 2 and 20 filters. Each output feature is obtained by convolving the corresponding elements in the previous layer's features with different convolutional kernels, adding the bias, and activating the result using the ReLU function. The pooling layer uses the max-pooling method, with a downsampling scale of 2×1. The convolution kernel weights are randomly initialized, and the biases are initialized to zero.

\begin{figure}[h!]
    \centering
    \captionsetup{font={small,bf}} 
    \includegraphics[width=0.9\textwidth]{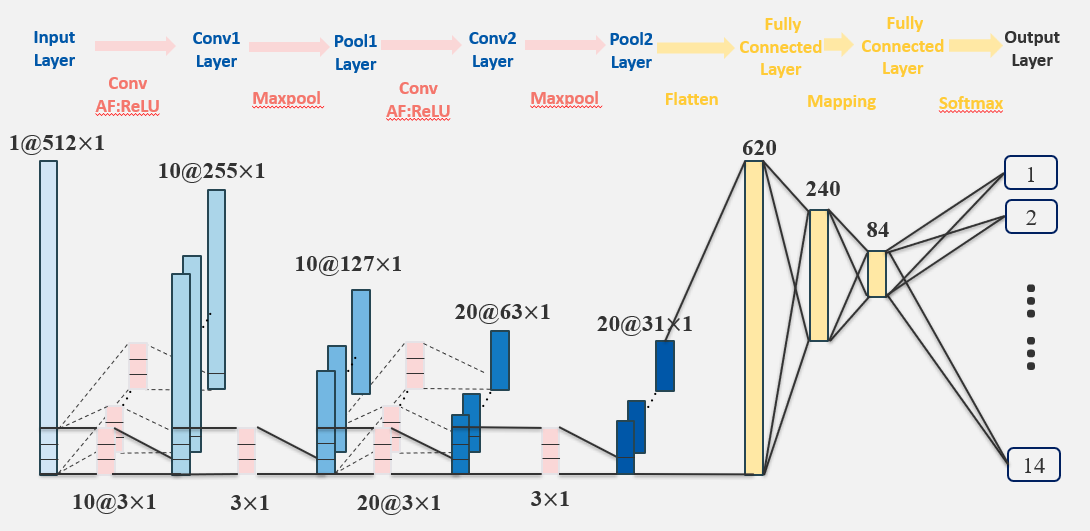}
    \caption{The architecture of CNN model}
    \label{fig3}
\end{figure}

In order to address the potential overfitting problem during network training, this paper incorporates the L2 regularization method and stochastic gradient descent with momentum (SGD with momentum) to enhance the model's generalization ability and computational efficiency. Shi B[14] provides a detailed discussion of the advantages of SGD with momentum, as well as the appropriate value of the momentum coefficient. In this study, the momentum coefficient is set to 0.9.

Hyperparameter optimization is performed using grid search, and the best-performing hyperparameters are as follows: the learning rate is set to 0.007, the regularization factor is set to 0.005, and the maximum number of iterations is 80. The batch size used for training is 35. The loss function curve and the test set accuracy curve obtained from training with this model are shown in Figure 4. The shaded area represents the range of loss function variations across different mini-batches in each epoch. It can be observed that with the increase in the number of iterations, issues of vanishing and exploding gradients emerge.

\begin{figure}[h!]
    \centering
    \captionsetup{font={small,bf}} 
    \includegraphics[width=0.7\textwidth]{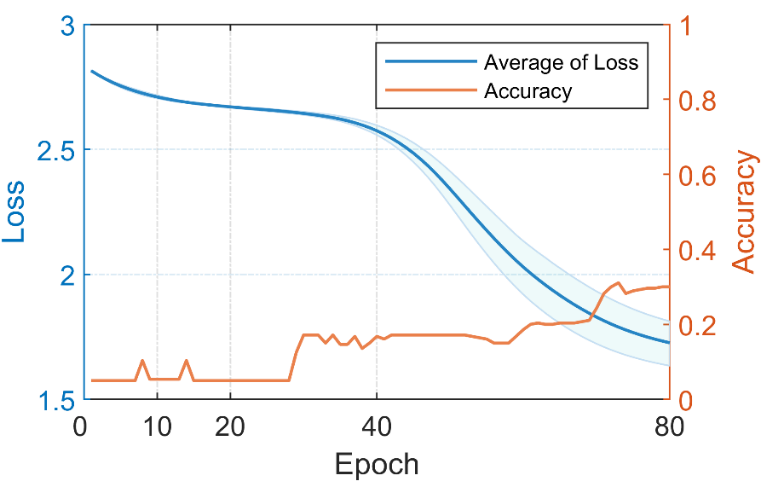}
    \caption{The accuracy and loss value during model training}
    \label{fig4}
\end{figure}

The results above indicate that classification using only CNN performs poorly. This is because CNN contains multiple convolutional and pooling layers, which require a large number of parameters, and the cost of parameter optimization and adjustment is high. Additionally, CNN suffers from issues such as vanishing and exploding gradients, and the long iteration times limit classification accuracy, resulting in a final classification accuracy of only 30.00\%. To further improve classification accuracy, this paper utilizes CNN as a feature extractor and feeds the 84 features extracted by the fully connected layer into a parameter-optimized SVM classifier for classification, thereby improving classification precision.

\subsection{Classifier: Support Vector Machine}
Venkatesan[15] used Support Vector Machines (SVM) for image classification and achieved excellent results. Based on their experiments, SVM outperformed other machine learning classifiers. The SVM constructs a hyperplane that maximizes the margin between the positive and negative classes.

First, in a binary classification problem, assuming that the data is linearly separable, the SVM aims to maximize the linear margin between two classes. The condition for linear separability is to identify an optimal hyperplane that not only correctly classifies the data but also maximizes the margin between classes. The equation of the hyperplane is defined as follows:

\begin{equation}
g(x) = \omega^T x + b
\end{equation}

In Equation (1), $\omega$ is the normal vector of the hyperplane, and $b$ is the bias (intercept) term.

However, the problem here is not always linearly separable. To handle such cases, a penalty parameter \(C\) and slack variables \(\xi_i\) are introduced to allow certain classification errors. Additionally, the Radial Basis Function (RBF) kernel is used to project the original data into a higher-dimensional feature space to address nonlinear classification problems. For \(N\) samples, the objective function is as follows:

\begin{equation}
\min \frac{1}{2} \|\omega\|^2 + C \sum_{i=1}^N \xi_i \quad \text{s.t.} \quad y_i (\omega^T x_i + b) + \xi_i - 1 \geq 0, \; \xi_i \geq 0, \; i = 1, 2, \ldots, N
\end{equation}

In Equation (2), \(\xi_i \geq 0\) are the slack variables, which measure the distance of misclassified samples from the corresponding hyperplane. The slack variables allow some samples to violate the margin requirement.

The kernel function used is the Radial Basis Function (RBF), which is defined as:

\begin{equation}
k(x_i, x) = \exp \left( -\frac{\|x_i - x\|^2}{\sigma^2} \right)
\end{equation}

Thus, the equation of the hyperplane can be expressed as:

\begin{equation}
f(x) = \omega^T x + b = \sum_{i=1}^N \alpha_i y_i k(x_i, x) + b
\end{equation}

In Equation (4), the Lagrange multipliers \(\alpha_i\) are obtained by solving the dual problem and take values within the range \(0 \leq \alpha_i \leq C\).

Finally, this study uses the \textit{one-vs-one} method to achieve multi-class classification. The 14 classes of data are paired for binary classification, and each support vector machine (SVM) is trained on two different classes. During the classification phase, a voting scheme is adopted to determine the final classification result. In the case of a tie, the sum of the absolute probability values is used as the criterion for the final classification output.

In the proposed model, the penalty parameter \(C\) and the kernel function parameter \(\sigma\) determine the performance of the support vector machine. If the penalty parameter \(C\) is too small, it may lead to underfitting; if it is too large, it may cause overfitting. \(\sigma\) is an important parameter of the kernel function, which controls the complexity of the model. Therefore, the choice of parameters plays a crucial role in both the efficiency and accuracy of classification. In this study, an improved particle swarm optimization (PSO) algorithm is adopted to optimize the parameters.
 
\subsection{Parameter Optimizer: Particle Swarm Optimization}

The Particle Swarm Optimization (PSO) algorithm originates from studies of bird foraging behavior. PSO is initialized as a swarm of randomly distributed particles, where each particle has velocity and position attributes. The velocity represents how quickly it moves, and the position determines the search direction in space. Each particle independently searches for the optimal solution in the search space and marks its current position as the individual best, \( pbest \). The particles share their individual best positions to determine the global best solution, \( gbest \). At each iteration, particles update their velocities and positions according to the formulas below:

\begin{equation}
\begin{cases}
v_i^t = \omega v_i^t + c_1 \text{rand()} (pbest_i - x_i) + c_2 \text{rand()} (gbest_i - x_i) \\
x_i^{t+1} = x_i^t + v_i^t
\end{cases}
\tag{5}
\end{equation}

In Equation (5), \( v_i \) represents the velocity of the \( i \)-th particle, and \( x_i \) represents the position of the \( i \)-th particle. \( pbest_i \) and \( gbest_i \) are the individual and global best solutions of the \( i \)-th particle, respectively, and \( \text{rand()} \) generates random numbers between \( (0,1) \).

The inertia weight \( \omega \) balances global exploration and local exploitation. A larger \( \omega \) is needed in the early stage when steps are larger, while a smaller \( \omega \) is required in later stages for stronger local development capabilities. To improve efficiency and avoid premature convergence to local optima, this study adopts an inertia weight decay strategy, where \( \omega \) decreases linearly according to Equation (6). Based on empirical values, \( \omega \) ranges from \( (0.4, 0.9) \). In this study, \( \omega \) starts at \( 0.9 \) and ends at \( 0.4 \):

\begin{equation}
\omega = 0.9 - 0.5 \frac{k}{k_{\text{max}}}
\tag{6}
\end{equation}

In Equation (6), \( k \) and \( k_{\text{max}} \) represent the current iteration number and the maximum iteration number, respectively.

The coefficients \( c_1 \) and \( c_2 \) are positive learning factors. Martinez[18] pointed out that setting \( c_1 = c_2 \) maximizes the stability region of the search. Additionally, assigning equal weights to \( pbest \) and \( gbest \) prevents the algorithm from prematurely falling into local optima. Following Jiang \textit{et al.}, this study adopts \( c_1 = c_2 = 1.5 \), with a population size of \( 40 \) and a maximum iteration number of \( 100 \).

\section{CPSVM Model}

This paper combines the characteristics of CNN feature extraction, SVM fast classification, and IPSO global optimization, and proposes a CNN and IPSO optimized SVM model—CPSVM. The model framework is shown in Figure 5.

\begin{figure}[h!]
    \centering
    \captionsetup{font={small,bf}} 
    \includegraphics[width=1.0\textwidth]{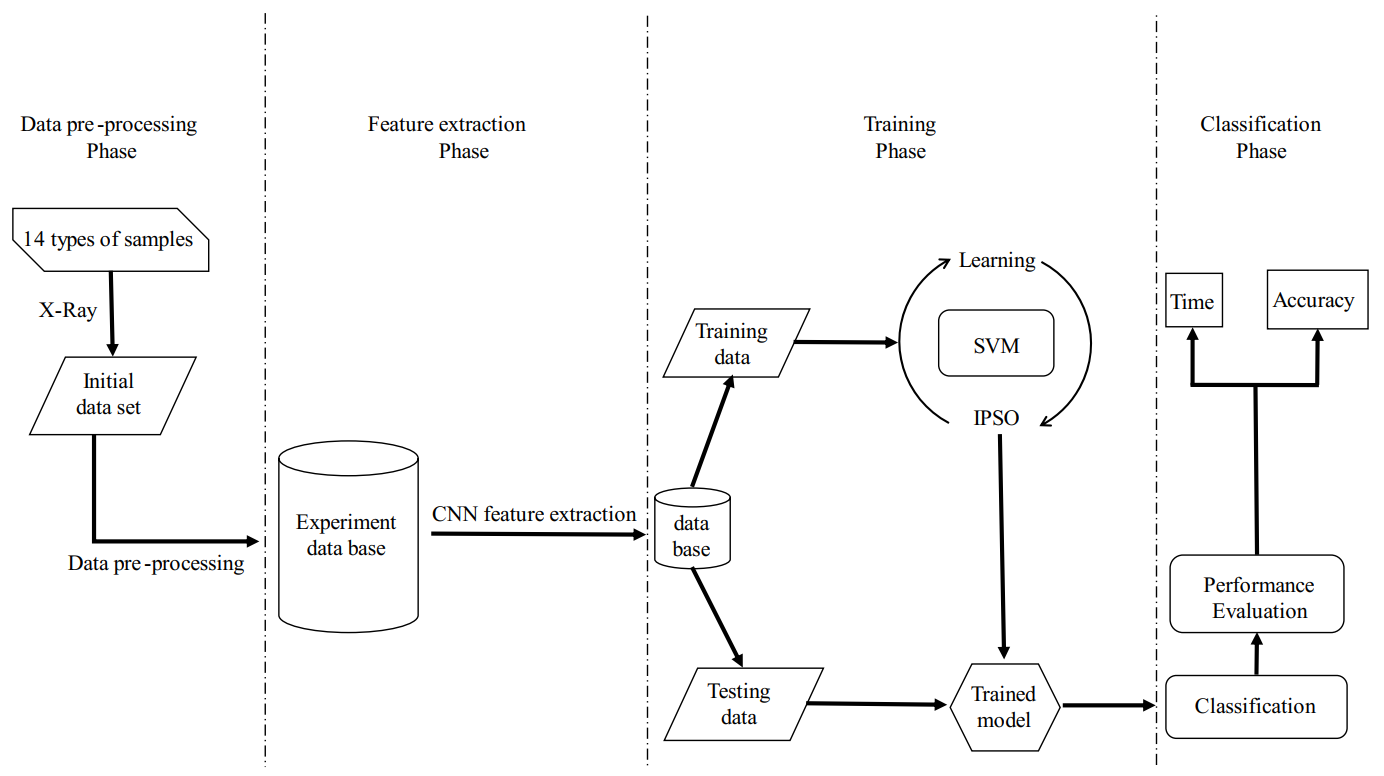}
    \caption{The Model Framework Diagram of CPSVM}
    \label{fig5}
\end{figure}

The specific implementation steps are as follows:\\
\textbf{Step 01:} Spectral data preprocessing. \\
\textbf{Step 02:} Label the processed data and shuffle the order. Randomly select 80\% of the processed data as the training set, with the remaining 20\% as the test set. \\
\textbf{Step 03:} Input the preprocessed data into the trained convolutional neural network and output 84 features from the last fully connected layer. \\
\textbf{Step 04:} Randomly initialize the population positions. Treat the vector $(C, \sigma)$ as a particle, and randomly initialize the penalty parameter $C$ and the kernel function parameter $\sigma$ for each support vector machine. \\
\textbf{Step 05:} Input the features obtained in Step 03 into the support vector machine and use this data to train the support vector machine. \\
\textbf{Step 06:} Input the test set data into the support vector machine trained in Step 05 to obtain the prediction accuracy. \\
\textbf{Step 07:} Define the fitness function of the IPSO algorithm as the prediction accuracy obtained in Step 06. \\
\textbf{Step 08:} Calculate the fitness value of each particle based on the fitness function. If the current fitness value is greater than $p_{\text{best}}$, replace the original $p_{\text{best}}$; otherwise, it remains unchanged. \\
\textbf{Step 09:} Compare the maximum $p_{\text{best}}$ with $g_{\text{best}}$. If $p_{\text{best}} > g_{\text{best}}$, replace $g_{\text{best}}$ with $p_{\text{best}}$; otherwise, retain $g_{\text{best}}$. \\
\textbf{Step 10:} Update the velocity and position of each particle according to equation (5). \\
\textbf{Step 11:} Check if the iteration count has reached the maximum value. If yes, proceed to Step 12; otherwise, increase the iteration count by 1 and go back to Step 05. \\
\textbf{Step 12:} Construct the SVM model based on the optimized parameters, classify the test set data extracted by CNN, and output the classification results.

The model flowchart is shown in Figure 6.

\begin{figure}[h!]
    \centering
    \captionsetup{font={small,bf}} 
    \includegraphics[width=0.8\textwidth]{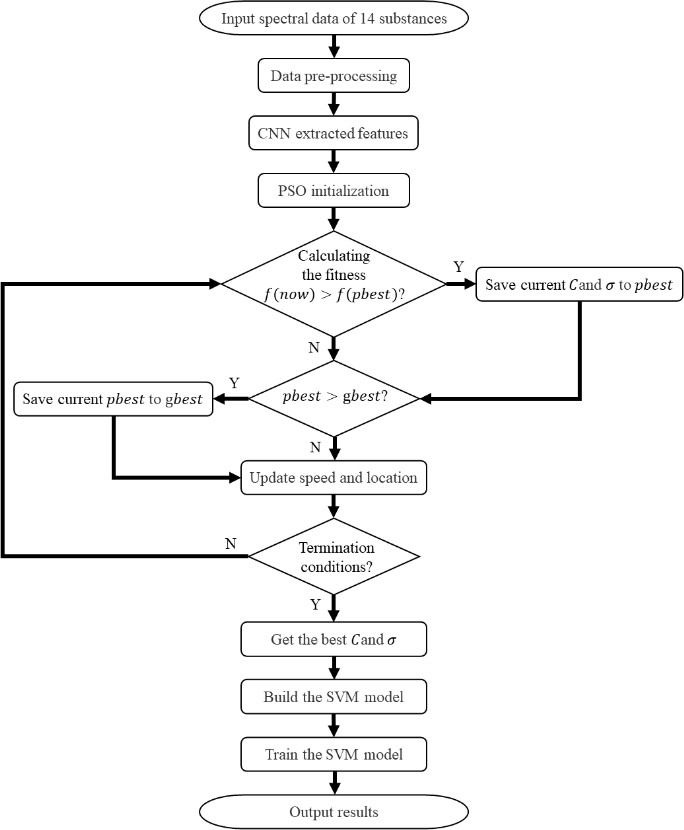}
    \caption{The Flow chart of CPSVM}
    \label{fig5}
\end{figure}

\section{Experimental Results and Analyses}

To evaluate the effectiveness of the models, we selected three models for comparison: the SVM model with randomly initialized parameters \( C \) and \( \sigma \), the IPSO-SVM model without feature extraction, and the CPSVM model utilizing CNN for feature extraction. The classification results of these models were compared and evaluated.

We constructed the normalized confusion matrices for the three models to determine the degree of confusion between different substances, as shown in Figure~7. The normalized confusion matrix indicates the proportion of correctly classified and misclassified samples for each category. From the normalized confusion matrices, we observed that all three models could distinguish the 14 substances. The SVM model with randomly initialized parameters \( C \) and \( \sigma \) exhibited the lowest classification accuracy, with a higher proportion of misclassifications. In contrast, the IPSO-SVM model and the CPSVM model achieved the highest recognition accuracy, with only a few instances of misclassification.

\begin{figure}[h!]
    \centering
    \captionsetup{font={small,bf}} 
    \includegraphics[width=0.45\textwidth]{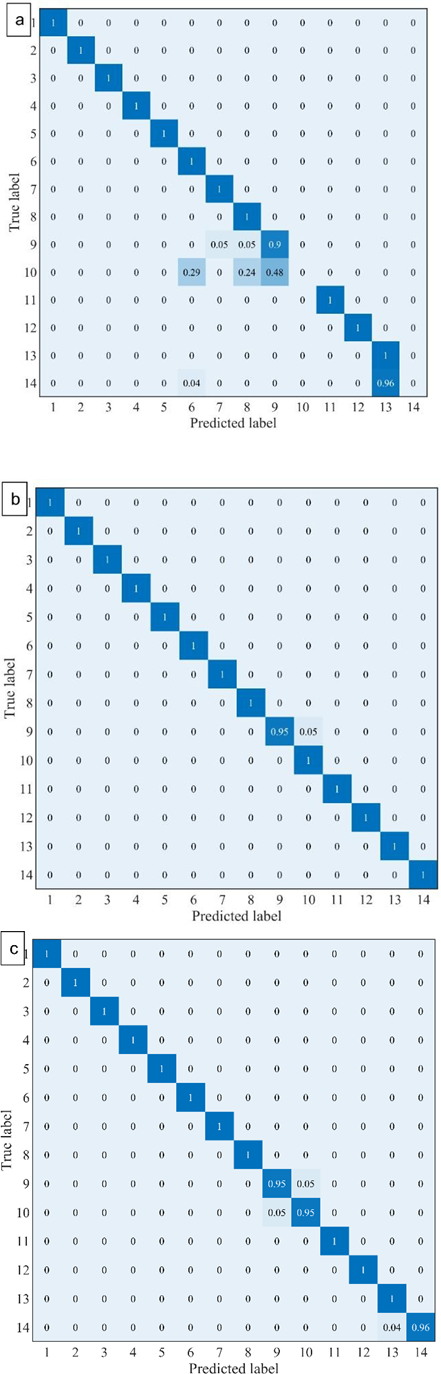}
    \caption{Normalized confusion matrix for (a) SVM(b) IPSO-SVM(c) CPSVM}
    \label{fig7}
\end{figure}

To visualize the process of particle swarm optimization in searching for optimal parameters, three-dimensional accuracy surface plots for the two models were generated, as shown in Figure~8. The IPSO-SVM model achieved a result of 99.6429\% with a runtime of 3694.6293 seconds, while the CPSVM model achieved a result of 99.2857\% with a runtime of 217.1137 seconds.

It can be observed that the runtime of the CPSVM model is nearly 17 times shorter than that of the IPSO-SVM model, while the prediction accuracy remains almost the same. This indicates that the CPSVM model can significantly improve efficiency without compromising prediction accuracy.

\begin{figure}[h!]
    \centering
    \captionsetup{font={small,bf}} 
    \includegraphics[width=0.8\textwidth]{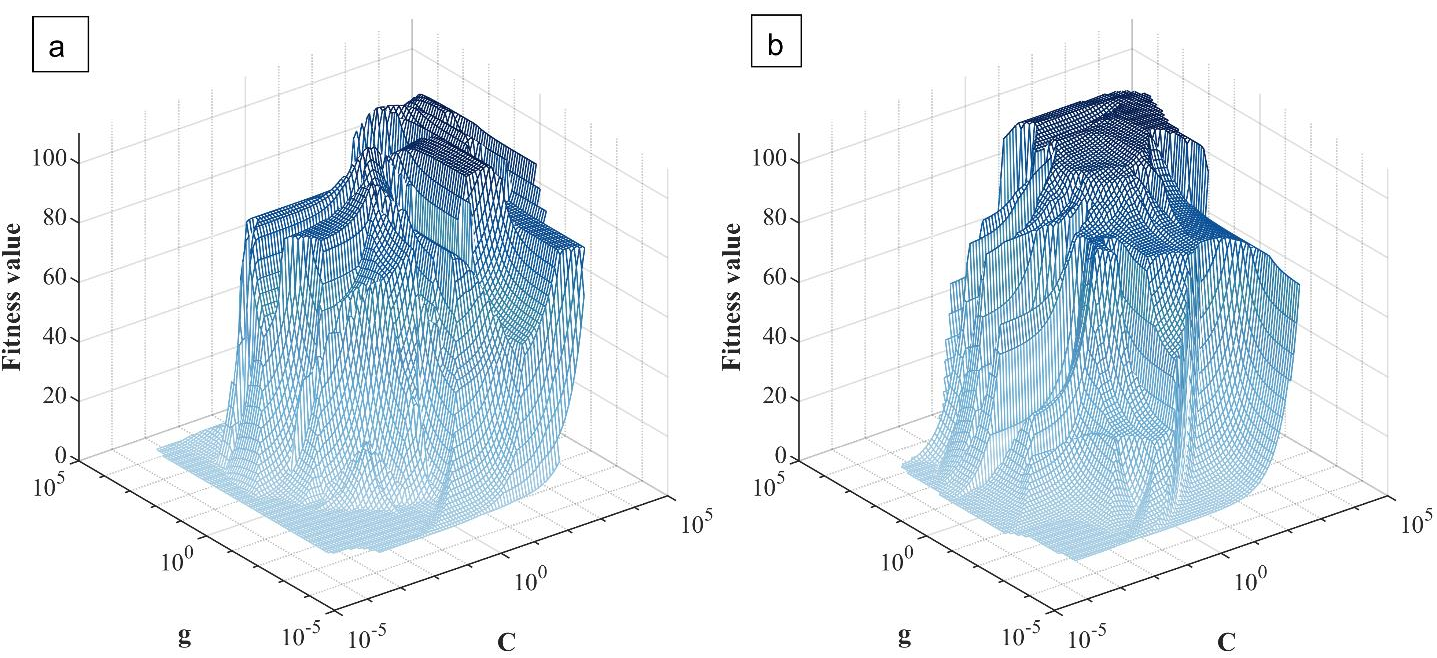}
    \caption{3D accuracy surface of two models (a) IPSO-SVM (b) CPSVM}
    \label{fig8}
\end{figure}

The fitness variation curves of IPSO-SVM and CPSVM during the first 10 iterations are shown in Figure~9. From the figure, it can be observed that the fitness values of the two models reach 99.6429\% and 99.2857\%, respectively. This indicates that the training process has reached a balanced state, and the parameter settings for the models are appropriate, achieving ideal and stable classification performance.

The average fitness curve represents the average fitness of the best positions of all particles during the evolutionary process. From the figure, it can be seen that the average fitness curve of CPSVM rises faster and reaches a higher value, indicating that the particles converge toward the optimal point more quickly, and the particle swarm exhibits faster convergence. The best fitness curve represents the overall fitness of the swarm, which remains stable throughout the evolutionary process, reflecting the stability of the final results.

\begin{figure}[h!]
    \centering
    \captionsetup{font={small,bf}} 
    \includegraphics[width=0.55\textwidth]{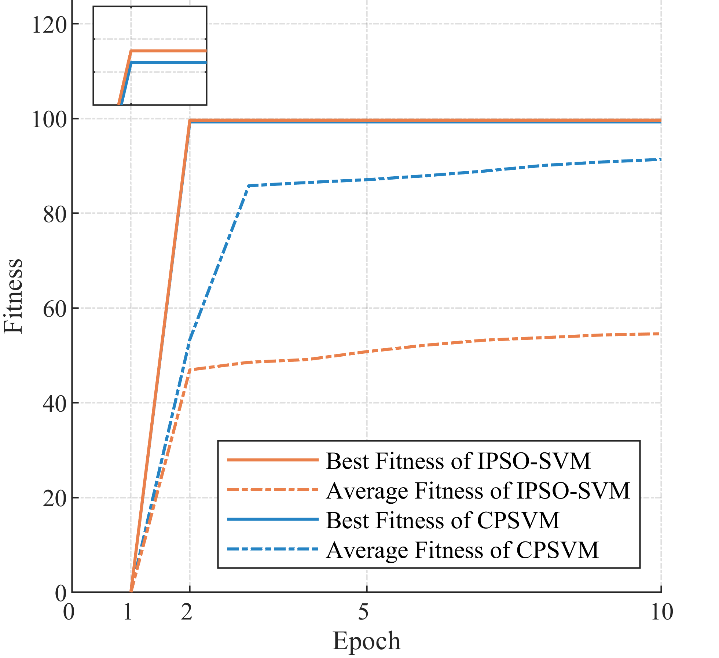}
    \caption{Fitness curves of IPSO-SVM and CPSVM}
    \label{fig9}
\end{figure}

Finally, the classification performance of different models over 20 Monte Carlo experiments was compared, as shown in Table~2. The results demonstrate that both improved models achieved excellent classification results for drug analogs, with classification accuracy exceeding 90\%.

It is noteworthy that the runtime of the improved algorithms is longer than that of the single SVM algorithm. This is because the improved models perform parameter optimization based on the SVM model. However, the trade-off in runtime yields significant benefits. The improved models can substantially enhance prediction accuracy while maintaining stable performance.

By comparing with the traditional IPSO-SVM, it can be observed that CPSVM performs better overall and is more stable. CPSVM significantly reduces the runtime while achieving similar prediction accuracy, enabling fast and accurate classification of different drug analogs.

\begin{table}[h]
\centering
\captionsetup{font={small,bf}} 
\caption{Comparison of the results of the three models}
\begin{tabular}{@{\hskip1cm}ccc@{\hskip1cm}}
\toprule
Model & Prediction Accuracy & Training Time \\
\midrule
SVM & $78.07 \pm 15.72$ & $4.42 \pm 1.77$ \\
IPSO-SVM & $99.64 \pm 0$ & $3694.63 \pm 736.76$ \\
CPSVM & $99.14 \pm 0.18$ & $217.11 \pm 7.28$ \\
\bottomrule
\end{tabular}
\end{table}

\section{Discussion}

In this study, the practical problem we aim to solve is the real-time detection of drugs in customs, airports, and similar locations. To address this, we first simplify the problem to the classification of drugs and evaluate the proposed methods based on two dimensions: classification accuracy and algorithmic complexity (i.e., runtime).

First, we selected 14 chemical reagents with similar chemical formulas to drugs, as detailed in Table~1. Using an X-ray absorption spectrometer in the laboratory, we obtained their X-ray absorption spectra. During preprocessing, the spectral images were compressed into one-dimensional data, which significantly improved the computational efficiency of the algorithms.

To achieve efficient detection of drug analogs, we developed a novel and simple method: a convolutional feature extraction and PSO-optimized SVM classification scheme. In this model, CNN is used to extract features from spectral data, SVM classifies the drugs based on the extracted features, and PSO optimizes the two hyperparameters of SVM to improve classification accuracy.

Experimental results demonstrate that the proposed model resolves the instability of results produced by the standalone SVM algorithm and the long runtime issue associated with PSO-SVM. Additionally, it avoids the gradient vanishing problem encountered when using CNN alone, thereby reducing the cost of parameter optimization and adjustment. The model exhibits satisfactory computational efficiency and accuracy, enabling effective classification of 14 drug analogs.

In the future, to make the experimental results more convincing, we will use the Raman spectra of drugs as the sample dataset. By ensuring a single-variable condition, various methods will be employed for classification, aiming to demonstrate the superiority of the X-ray absorption spectroscopy method. We will also explore additional methods, including reagent detection, and compare the advantages and disadvantages of these methods in terms of detection time and accuracy.


\begin{thebibliography}{99}

\bibitem{ref1} Dies H, Raveendran J, et al. Sensors and Actuators B-Chemical, 2018, 257: 382--388.

\bibitem{ref2} Dong R, Weng S, Yang L, et al. Anal Chem, 2015, 87(5): 2937--2944.

\bibitem{ref3} Fang Z, Wang M, Hu W, et al. Computers and Electronics in Agriculture, 2021, 183: 106062.

\bibitem{ref4} WANG Qian, YANG Zheng, FANG Zheng, et al. (Qian Wang, Zheng Yang, Zheng Fang). Chemical Journal of Chinese Universities, 2018, 39(07): 1434--1439.

\bibitem{ref5} Xue Z H, Du P J, Su H J. IEEE Journal of Selected Topics in Applied Earth Observations and Remote Sensing, 2014, 7(6): 2131--2146.

\bibitem{ref6} Kennedy J, Eberhart R. Proceedings of ICNN'95-International Conference on Neural Networks, 1995.

\bibitem{ref7} Sha X Y, Fang G Q, Cao G X, et al. Analyst, 2022, 147(24): 5785--5795.

\bibitem{ref8} Liu L T, Li X Y. Digital Signal Processing: A Review Journal, 2022, 123.

\bibitem{ref9} Shang H, Shang L W, Wu J J, et al. Spectrochimica Acta Part A: Molecular and Biomolecular Spectroscopy, 2023, 287.

\bibitem{ref10} Chen J S, W P, Tian Y B, et al. Journal of Biophotonics, 2023, 16(2).

\bibitem{ref11} Zhang J H, Zhang J M, Ding J Y, et al. Vibrational Spectroscopy, 2022, 118.

\bibitem{ref12} Laeli A R, Rustam Z, Pandelaki J, et al. 2021 International Conference on Decision Aid Sciences and Application (DASA). IEEE, 2021: 682--686.

\bibitem{ref13} Hu B, Zhang X L, Ouyang Q N, et al. Measurement, 2016, 93: 252--257.

\bibitem{ref14} Shi B. On the Hyperparameters in Stochastic Gradient Descent with Momentum. ArXiv, 2021.

\bibitem{ref15} Venkatesan C, Karthigaikumar P, Paul A, et al. IEEE Access, 2018, 6: 9767--9773.

\bibitem{ref16} Eberhart, Shi Y H. Proceedings of the 2001 Congress on Evolutionary Computation, 2001, 1: 81--86.

\bibitem{ref17} Martinez J L F, Gonzalo E G. Swarm Intelligence, 2009, 3(4): 245--273.

\bibitem{ref18} Zhen Yuan, Qizhi Zhang, E. Sobel, Huabei Jiang.Comparison of diffusion approximation and higher order diffusion equations for optical tomography of osteoarthritis[J]. Biomed. Opt. Express, 2009, 14 (5):13-27.

\end{thebibliography}
\end{document}